\documentclass{bmcart}

\usepackage[utf8]{inputenc} 

\def\includegraphics{}

\startlocaldefs
\endlocaldefs

\usepackage{latexsym,bm}
\usepackage{graphics}
\usepackage{subfigure}
\usepackage{amssymb,amsmath}
\usepackage[normalem]{ulem}
\usepackage{graphicx}
\usepackage{url}
\usepackage{lineno,hyperref}

\usepackage{multirow}

\begin{document}

\begin{frontmatter}

\begin{fmbox}
\dochead{Research}


\title{Multi-stage Multi-task feature learning via adaptive threshold}


\author[
   addressref={aff1},                   
   ]{\inits{YF}\fnm{Yaru} \snm{ Fan}}
\author[
   addressref={aff1,aff2,aff3},
   corref={aff1},  
   email={yilun.wang@rice.edu}
]{\inits{YW}\fnm{Yilun} \snm{Wang}}
\author[
   addressref={aff1},                   
   ]{\inits{TH}\fnm{Tingzhu} \snm{ Huang}}

\address[id=aff1]{
  \orgname{School of Mathematical Sciences, University of Electronic Science and Technology of China}, 
  \street{Xiyuan Ave},                     %
  \postcode{611731}                                
  \city{Chengdu},                              
  \cny{China}                                    
}
\address[id=aff2]{
  \orgname{Center for Information in Biomedicine, University of Electronic Science and Technology of China}, 
  \street{Xiyuan Ave},                     %
  \postcode{611054}                                
  \city{Chengdu},                              
  \cny{China}                                    
}
\address[id=aff3]{
  \orgname{Center for Applied Mathematics, Cornell University}, 
  \street{Xiyuan Ave},                     %
  \postcode{611731}                                
  \city{Chengdu},                              
  \cny{China}                                    
}

\begin{artnotes}
\end{artnotes}

\end{fmbox}


\begin{abstractbox}

\begin{abstract} 
Multi-task feature learning aims to identity the shared features among tasks to improve generalization. It has been  shown that by minimizing  non-convex learning models,  a better solution than the convex alternatives can be obtained.  Therefore, a non-convex model based on the capped-$\ell_{1},\ell_{1}$ regularization was proposed in \cite{Gong2013}, and a corresponding efficient multi-stage multi-task feature learning algorithm (MSMTFL) was presented.  However, this algorithm harnesses a prescribed fixed threshold in the definition of the capped-$\ell_{1},\ell_{1}$ regularization and the lack of adaptivity might result in suboptimal performance. In this paper we propose to employ an adaptive threshold in the capped-$\ell_{1},\ell_{1}$ regularized formulation, where the corresponding  variant of MSMTFL will incorporate an additional component to adaptively determine the threshold value. This variant is expected to achieve  a better feature selection performance over  the original MSMTFL algorithm. In particular, the embedded adaptive threshold component comes from our previously proposed iterative support detection (ISD) method \cite{Wang2010}. 
Empirical studies on both synthetic and real-world data sets demonstrate the effectiveness of this new variant over the original MSMTFL.
\end{abstract}




\end{abstractbox}
%

\end{frontmatter}



\section{Introduction}
A fundamental limitation of the common machine learning methods is the cost incurred by the preparation of the large training samples required for good generalization. Multi-task learning (MTL) offers a potential remedy. Unlike common single task learning, MTL accomplishes tasks simultaneously with other related tasks, using a shared representation. One general assumption of multi-task learning is that all tasks should share some common structures, including a similarity metric matrix \cite{Ghosn1997}, a low rank subspace \cite{Chen2010, Negahban2011}, parameters of Bayesian models \cite{Yu2005} or a common set of features \cite{Argyriou2008, Kim2009, Lounici2009}.  Improved generalization is achieved because  what is learned from each task can help with the learning of other tasks \cite{Gong14}. MTL has been successfully applied to many applications such as stock selection \cite{Ghosn1997}, speech classification \cite{Parameswaran2010} and medical diagnoses \cite{Bi2008}.

While the majority of  existing  multi-task feature learning algorithms assume that the relevant features are shared by all tasks, some studies have begun to consider a more general case where  features can be  commonly shared only among most, but not necessarily all of them. In other word,  they try to learn the features specific  to each task
as well as the common features shared among tasks \cite{Gong2013}.  In addition, MTL  is commonly  formulated as a convex regularization problem. Thus the resultant models are restrictive and suboptimal. In order to remedy the above two shortcomings, a specific non-convex formulation based on the capped-$\ell_{1},\ell_{1}$ regularized formulation for multi-task sparse feature learning, was proposed in \cite{Gong2013}. 
Then a corresponding  Multi-Stage Multi-Task Feature Learning (MSMTFL) algorithm is presented.  This non-convex model and corresponding algorithm usually achieves a better solution than the corresponding convex models. However, like many non-convex algorithms, it may not get a globally optimal solution which is often computationally prohibitive to obtain in practice.

Notice that the above capped-$\ell_{1},\ell_{1}$ regularized formulation employs a prescribed fixed threshold value in its definition, and the MSMTFL employs the same fixed threshold value.  This threshold value is used to determine the large rows of the unknown weight matrix. However, the result is difficult to  prescribe beforehand, because this value is data-dependent. Thus a natural idea is to adaptively determine it in practice, in order to achieve even better performance. In this paper, we propose to incorporate  an adaptive threshold estimation scheme into the MSMTFL.  In particular,  we consider the structure information of the intermediate estimated solution of the current stage and make use of this information to estimate the threshold value, which will be used in the next stage. This way, the threshold value and the resultant solution will be updated in an alternative way.  As for the adopted threshold value estimation scheme, we are motivated by the iterative support detection method first proposed by Wang et al \cite{Wang2010} which employed the ``first significant jump" heuristic rule to estimate the adaptive threshold value.  We leverage the rule to the MSMTFL in order to solve the non-convex feature learning formulation with the adaptive threshold, though other methods of determining the adaptive threshold could be applied. With the help of the adaptive threshold value, rather than a prescribed fixed value, an even better feature learning result is expected to be  obtained for both synthetic and real data tests. 

\textbf{Organization}: First, we will  review the original MSMTFL algorithm in  section $2$. In section $3$, we will present the method to  adaptively set the threshold in MSMTFL.  In section $4$, several numerical experiments  verify the effectiveness of our new algorithm. In the conclusion we will describe future work. 

\textbf{Notation}: Scalars and vectors are denoted by lower case letters. Matrices and sets are denoted by capital letters. The Euclidean norm, $\ell_{1}$ norm, $\ell_{\infty}$ norm, $\ell_{p}$ norm, $\ell_{0}$ norm and Frobenius norm are denoted by $\|\cdot\|$,$\|\cdot\|_{1}$, $\|\cdot\|_{\infty}$, $\|\cdot\|_{p}$, $\|\cdot\|_{0}$ and $\|\cdot\|_{F}$, respectively. $w_{i}$ and $w^{j}$ denote the $i$-th column and $j$-th row of matrix $W$, respectively. $|\cdot|$ denotes the absolute value of a scalar or the number of elements in a set. Then we define $N_{m}$ as the set $\{1,\cdots, m\}$ and $N(\mu, \sigma^{2})$ as a Gaussian distribution with mean $\mu$ and standard deviation $\sigma$.

\section{Review of the MSMTFL Algorithm} \label{Sec:MSMTFL}
In this section, we first give a brief introduction to multi-task feature learning (MTFL) exploiting the shared features among tasks. Then we  review the capped-$\ell_{1},\ell_{1}$ regularized feature learning model and corresponding  MSMTFL algorithm proposed in \cite{Gong2013}.

Consider $m$ learning tasks associated with the given training data $\{(X_{1}, y_{1})$,$\cdots$, $(X_{m}, y_{m})\}$, where $X_{i}\in R^{n_{i}\times d}$ is the data matrix of the $i-$th task with each row being a sample, $y_{i}\in R^{n_{i}}$ is the response of the $i-$th task, and $n_{i}$ is the number of samples for the $i-$th task. Each column of $X_{i}$ represents a feature and $d$ is the number of features. In the scenario of MTL, it is assumed that the different tasks  share the same set of features, i.e. the columns of different $X_{i}$   represent the same features.  The purpose of  feature selection is to learn a weight matrix $W=[w_{1},\cdots, w_{m}]\in R^{d\times m}$, consisting of the weight vectors for $m$ linear predictors: $y_{i}\approx f_{i}(X_{i})= X_{i}w_{i}$, $i\in N_{m}$. The quality of prediction is measured through a loss function $l(W)$ and we assume that $l(W)$ is convex in $W$ throughout the paper. For each $X_i$, the magnitude of the components of $w_i$ reflects the importance of the corresponding column (feature) of $X_i$. In MTL, the important features among different tasks are assumed to be almost the same and  should be used when establishing an MTL model.

We consider MTL in the scenario that $d\gg m$, i.e there are many more features than tasks, because this situation  commonly arises in many applications.  In such cases, an unconstrained empirical risk minimization is inadequate to obtain a reliable solution because overfitting of the training data is likely to occur. A standard remedy for this problem is to impose a constraint on $W$
to obtain a regularized feature learning model.  This constraint or regularization reflects our prior knowledge of the desired solution. For example, we have learned that one important aspect of $W$ is sparsity, because only a small portion of the features are really relevant to the given tasks. It is well known that the $L_{0}$ regularization is a natural choice for the sparsity regularization. If  the sparsity parameter $k$ for the target vector is known,  a feature learning model based on $L_{0}$ regularization is as follows:
\begin{equation} \label{eq:1}
 \widehat{W}= \arg \min_{W} l(W) \qquad s.t. \qquad ||w_{i}||_{0}\leq k \quad i=1, \ldots, m.
\end{equation}
where $l(W)$ is usually the quadratic loss function of $W$, i.e.
$
l(W)= \sum_{i=1}^{m} \frac{1}{mn_{i}} \|X_{i}w_{i}-y_{i}\|^{2}.
$ 
 However, the sparsity parameter $k$ is often unknown in practice and the following unconstrained formulation can be considered instead. 
\begin{equation} \label{eq:2}
\widehat{W}= \arg \min_{W} \{l(W)+ \lambda g(W) \}
\end{equation}
where $g(W) \doteq \sum_{i=1}^{m} ||w_{i}||_{0}$ is the sparsity regularization term and $\lambda (> 0)$ is a parameter balancing the empirical loss and the  regularization term.  
However,  either (\ref{eq:1}) or (\ref{eq:2}) seems to  consider the sparsity of $w_i$ individually and fails to make use of the fact that different tasks share almost the same relevant features, i.e. the solution of MTL has a joint sparsity structure. In addition, another fundamental difficulty for this model is the prohibitive computational burden since the $L_{0}$ norm minimization is an NP-hard problem.

In order to make use of the joint sparsity structure of the solution of MTL,  ones often turn to the mixed $\ell_{p,q}$ norm based joint sparsity regularization \cite{Kowalski09mixed}. $$g(W) \doteq \|W\|_{p,q}= \sum_{j=1}^{d} (\|w^{j}\|_p^q)^{1/q}$$
where $w^j$ is the $j$-th row of $W$,   $p \ge 1$ and $ q \ge 0$. One common choice of $p, q$ is that $p=2$ and $q=1$ \cite{Yuan06L21}. It assumes that all these tasks share exactly the same set of the features, i.e. the relevant features to be shared by all tasks.

 However, it is too restrictive in real-world applications to require the relevant features to be shared
by all tasks. In order for  a
certain feature to be
shared by some  but not all tasks, many efforts have been made.  \cite{Jalali13dirtymodel} proposed to use $\ell_1$ + $\ell_{1, \infty}$ regularized formulation to  leverage the common features shared among tasks.  However, $\ell_1$ + $\ell_{1, \infty}$ regularizer is a convex relaxation of an $\ell_0$-type one, for example, $\ell_{1,0}$.  The convex regularizer
is known to be too loose to approximate the $\ell_0$-type one and often achieves suboptimal performance.
 %
  With $0 \le  q< 1$, the regularization $||W||_{p,q}$ is non-convex and expects to achieve a better solution theoretically. However, it is difficult to solve in practice. Moreover, the solution of the non-convex formulation heavily depends on the specific optimization algorithms employed and may result in  different solutions.

   In \cite{Gong2013},   a non-convex formulation, based on a capped-$\ell_{1}, \ell_{1}$
regularized model for multi-task feature learning is proposed, and   a corresponding Multi-Stage Multi-Task Feature
Learning (MSMTFL) algorithm is also given.  The capped-$\ell_{1}, \ell_{1}$ regularization is defined as  $g(W)= \sum_{j=1}^{d} \min(\|w^{j}\|_{1}, \theta)$. This proposed model aims to simultaneously learn the features specific to each task
 and the common features shared among tasks.  The advantage of MSMTFL is that the convergence, reproducibility analysis, and theoretical analysis for better performance over the convex models are given. In addition, the capped-$\ell_{1}, \ell_{1}$  regularization is a good approximation to $\ell_{0}$-type norm because as $\theta\rightarrow 0$, $\sum_{j=1} \min(\|w^{j}\|_{1}, \theta)/\theta\rightarrow ||W||_{1,0}$, which is  mostly preferred for its theoretically optimal sparsity enforcement. 


Specifically,  the model for multi-task feature learning with the capped-$\ell_{1}, \ell_{1}$ regularization is as follows:
\begin{equation}\label{eq:nonconvex}
\min_{W}\{l(W)+\lambda \sum_{j=1}^{d} \min(\|w^{j}\|_{1}, \theta)\},
\end{equation}
where $\theta (> 0)$ is a thresholding parameter, which distinguishes nonzeros and zero components; $w^{j}$ is the $j$-th row of the matrix $W$.
Obviously, due to the capped-$\ell_{1},\ell_{1}$  penalty, the optimal solution of problem (3) denoted as $W^{\star}$ has many zero rows. Due to the $\ell_{1}$ penalty on each row of $W$, some entries of the nonzero row  may be zero. Therefore, a certain feature can be shared by some  but not necessarily all the tasks under the formulation (3).

The multi-stage multi-task feature learning (MSMTFL) algorithm (see Algorithm $1$) based on the work \cite{Zhang2010,Zhang2012} is proposed in \cite{Gong2013} to solve problem (3). Note that for the first step of MSMTFL ($\ell=1$), the MSMTFL algorithm is equivalent to solving the $\ell_{1,1}$ regularized multi-task feature learning model (Lasso for MTL). Therefore, the final solution of the MSMTFL algorithm can be considered as a refinement of Lasso's for MTL.
Although MSMTFL may not find a globally optimal solution, 
 \cite{Gong2013} has theoretically shown that the solution obtained by Algorithm 1 improved the performance of the parameter estimation error bound as the multi-stage iteration proceeds, under certain circumstances. More details about intuitive interpretations, parameter settings and convergence analysis of the MSMTFL algorithm are provided in \cite{Gong2013}.  In addition, we need to point out that the key thresholding parameter $\theta$ is bounded below, that is $\theta \ge a m \lambda$, where $a$ is a constant. An exact solution  could be obtained from the Algorithm 1 with an appropriate thresholding parameter.
\\
\\
\begin{tabular}{l}
\hline
\textbf{Algorithm 1} The MSMTFL Algorithm  \\
\hline
1.Initialize $\lambda_{j}^{0}= \lambda$ ;\\
2.for $\ell= 1, 2,\cdots$,  do \\
  (a)$\quad$ Let $\hat{W}^{(\ell)}$ be a solution of the following problem:\\
     $\quad$ $\min_{W}$ $\{l(W)+ \sum_{j=1}^{d}\lambda_{j}^{(\ell-1)}\|w^{j}\|_{1}\}$.\\
  (b)$\quad$ Let $\lambda_{j}^{(\ell)}= \lambda I(\|(\hat{w}^{(\ell)})^{j}\|_{1}< \theta) (j=1,\cdots, d)$,\\
     $\quad$ where $(\hat{w}^{(\ell)})^{j}$ is the $j-$th row of $\hat{W}^{(\ell)}$ and $I(\cdot)$ denoted \\
     $\quad$ the $\{0,1\}$ valued indicator function.\\
   end. \\
\hline
\end{tabular}
\\
\section{Our Proposed Formulation and Algorithm}

In this paper, we would like to present a further study on  the capped-$\ell_{1},\ell_{1}$ regularization based multi-task feature learning model and corresponding MSMTFL algorithm. 

One limitation of the formulation (3) is to  employ a  fixed threshold $\theta$ to learn a weight matrix $W$. This prescribed value may be far from optimal because the choice of $\theta$ may vary greatly for different kinds of training data $X$ and difficult to  determine beforehand. In order to achieve a  better performance, we propose an empirical  heuristic method, which aims to learn  an adaptive threshold adopted in the non-convex multi-task feature learning formulation  (3). The key point is that we can refine threshold adaptively in each stage  according to the $W$ result of the last stage. In particular, we make use of  the ``first significant jump" rule proposed in \cite{Wang2010}, which automatically determines an appropriate threshold value of the current stage based on the recent $W$ result, though other possible  ways to adaptively determine the $\theta$ can also be applied.

Specifically, we propose to modify the original  non-convex multi-task feature learning formulation (\ref{eq:nonconvex}) with an unknown threshold value $\theta^{(W)}$, as follows. 
\begin{equation} \label{eq:AT}
\min_{W}\{l(W)+\lambda \sum_{j=1}^{d} \min(t_{[j]}, \theta^{(W)})\},
\end{equation}
where $l(W)$ is still the quadratic loss function as defined above; $\lambda(>0)$ is a parameter balancing the quadratic loss and the regularization; $t_{[j]}=\|w^{j}\|_{1}$. We can see that the difference between the models (\ref{eq:AT}) and (\ref{eq:nonconvex}) is the threshold parameter. In the new model, $\theta^{(W)}$ is data dependent and assumed to be unknown, rather than  a prescribed known value. While it seems to be natural idea, the  determination of the $\theta^{(W)}$ is not trivial and we will present an efficient adaptive method based on the "first significant jump" rule to estimate it. More details about ``first significant jump"  will be reviewed in Section \ref{dec:determineTheta}.

To solve the above optimization problem  (\ref{eq:AT}), we correspondingly modify the multi-stage multi-task feature learning algorithm proposed in \cite{Gong2013} by incorporating  an  adaptive threshold estimation step  (MSMTFL-AT) (see Algorithm 2 for more details).  At each stage (iteration, or step) in the Algorithm 2, the threshold $\theta^{(\ell)}$ will be updated according to the most recently learned solution $\hat{W}^{(\ell)}$.  $\theta^{(\ell)}$ keeps updating as iterations proceed and eventually reaches a stable and small range empirically.  
\\
\\
\begin{tabular}{l}
\hline
\textbf{Algorithm 2} The MSMTFL-AT Algorithm  \\
\hline
1.Initialize $\lambda_{j}^{0}= \lambda$ ;\\
2.for $\ell= 1, 2,\cdots$, do \\
  (a)$\quad$ $\hat{W}^{(\ell)}$ $\leftarrow$ $\min_{W} \{\ell(W)+ \sum_{j=1}^{d}\lambda_{j}^{(\ell-1)}t_{[j]}\}$;\\
  (b)$\quad$ $\theta^{(\ell)}$ $\leftarrow$ ``first significant jump" rule, using $\hat{W}^{(\ell)}$ as the reference;\\
  (c)$\quad$ $\lambda_{j}^{(\ell)}= \lambda I(\hat{t}^{(\ell)}_{[j]}< \theta^{(\ell)}) \qquad (j=1,\cdots, d)$, \\
    $\quad$ where $\hat{t}^{(\ell)}_{[j]}= \|(\hat{w}^{(\ell)})^{j}\|_{1}$ ($(\hat{w}^{(\ell)})^{j}$ is the $j$-th row of $\hat{W}^{(\ell)}$) \\
    $\quad$ and $I(\cdot)$ denoted the $\{0,1\}$ valued indicator function ;\\
  end. \\
\hline
\end{tabular}
\\

As shown in \cite{Gong2013}, the MSMTFL achieves a better estimation error bound than the convex alternative such as the $\ell_{1,1}$ regularized model, though  in general it only leads to a local critical point  for the non-convex problem (\ref{eq:nonconvex}). This enhancement is achieved because this local critical point  is a refinement based on the solution of the initial convex model, which is  the $\ell_{1,1}$ regularized model. Furthermore, our proposed  MSMTFL-AT is expected to achieve an even better estimation accuracy than the original MSMTFL because of the incorporated adaptive threshold estimation scheme. Here we only present an intuitive explanation, which is partially borrowed from the idea of the analysis of the iterative reweighted $\ell_2$ algorithm for $\ell_p$ ( $0 \le p < 1$ ) based non-convex compressive sensing in \cite{Chartrand08IRL2}.  We extended it from the common sparsity regularization to joint sparsity regularized considered in this paper. Note that the non-convex formulations (\ref{eq:nonconvex}) or (\ref{eq:AT}) based on capped-$\ell_{1},\ell_{1}$ regularization can be a good approximation to $L_{0}$ regularization if $\theta$ or $\theta^{(W)}$ is small, because  $\sum_{j=1}^{d} \min(t_{[j]}$, $\theta)/\theta$ is close to  $||t||_{0}$, where $t=[t_{[1]}, t_{[2]},  \ldots, t_{[d]}]$. 
Indeed, in practice, the $\theta^{(\ell)}$  of Algorithm 2  usually  decreases gradually as the iteration proceeds, though not necessarily always monotonically, due to the inherent estimation errors. Based on this observation,  an intuitive explanation why MSMTFL-AT could achieve a better performance than MSMTFL, is that an adaptive but relatively large $\theta^{(\ell)}$ at the beginning of MSMTFL-AT results in undesirable local minima being ``filled in". Once $\hat{W}^{(\ell)}$  is in the correct basin, decreasing $\theta^{(\ell)}$ allows the basin to deepen and $\hat{W}^{(\ell)}$ is expected to approach the true $W$ more closely. We expect to turn these notions into a rigorous proof in the future work.  In addition, suppose that $\lim_{\ell\rightarrow \infty}\theta^{(\ell)}$ exists and we denote it as $\theta^{(\bar{W})}$, then we intuitively expect that $\hat{W}^{(\ell)}$ of MSMTFL-AT converges to a critical point of the problem (\ref{eq:AT}) with $\theta^{(W)}$ being $\theta^{(\bar{W})}$. The rigorous proof will also constitute an important future research topic.



The reproducibility of the MSMTFL-AT algorithm can be guaranteed, as can the original MSMTFL. Theorem 2 in \cite{Gong2013} proves that the optimization problem 2(a) in MSMTFL (Algorithm 1) has a unique solution if $X_{i} \in R^{n_{i}\times d}$ has entries drawn from a continuous probability distribution on $R^{n_{i}d}$. Considering that  problem 2(a) in MSMTFL-AT (Algorithm 2) has the same formula as the one in MSMTFL,   the solution $\hat{W}^{(\ell)}$ of the MSMTFL-AT is also unique, for any $\ell \geq1$.    The main difference between MSMTFL-AT and  MSMTFL is  Step 2(b), which has a unique output. 
Therefore the solution generated by Algorithm 2 is also unique, since Steps 2(a), 2(b) and 2(c) all return unique results. 

\subsection{Adaptive Threshold in MSMTFL-AT} \label{dec:determineTheta}
While there could be many different rules for $\theta^{(\ell)}$, our choice is based on locating the ``first significant jump"  in the increasingly sorted sequence $|\hat{t}_{[j]}^{(\ell)}|$. For simplicity, after sorting, we still use the same notation,   where $|\hat{t}_{[j]}^{(\ell)}|$ denotes the $j$-th smallest among all  $\{\hat{t}_{[j]}^{(\ell)}\}_{j=1}^d$ by magnitude, as used in \cite{Wang2010}. Denote  $\hat{t}^{(\ell)}=[\hat{t}_{[1]}^{(\ell)}, \hat{t}_{[2]}^{(\ell)},  \ldots, \hat{t}_{[d]}^{(\ell)}]$. This rule looks for the smallest $j$ such that
\begin{equation}
|\hat{t}_{[j+1]}^{(\ell)}|- |\hat{t}_{[j]}^{(\ell)}|>\tau^{(\ell)}.
\end{equation}
This amounts to sweeping the increasing sequence $|\hat{t}_{[j]}^{(\ell+1)}|$ and looking for the first jump larger than $\tau^{(\ell)}$. Then we set $\theta^{(\ell)}$= $|\hat{t}_{[j]}^{(\ell)}|$. This rule has proved capable of  detecting many true nonzeros with few false alarms if the sequence $|\hat{t}_{[j]}^{(\ell)}|$ has the fast-decaying property \cite{Wang2010}.  In addition, several simple and heuristic methods have been adopted to define $\tau^{(\ell)}$ for different kinds of data matrix, as suggested in \cite{Wang2010}. In this paper, we adopt the following method for  following both synthetic and real data experiments, though other heuristic formulas could be proposed and tried. 
\begin{equation} \label{eq:tau}
\tau^{(\ell)}$= $n^{-1}\|\hat{t}^{(\ell)}\|_{\infty},
\end{equation} where $n=\sum_{i=1}^m n_i$ is the number of all samples.  An excessively large $\tau^{(\ell)}$ results in penalizing too many true nonzeros, while an excessively small $\tau^{(\ell)}$ results in ignoring too many false nonzeros and leads to low quality of solution. MSMTFL-AT will be quite effective with an appropriate $\tau^{(\ell)}$, though the proper range of $\tau^{(\ell)}$ might be case-dependent \cite{Wang2010}, due to the No Free Lunch theorem for the nonconvex optimization or learning \cite{WOLPERT1997}. However, numerical experiments have shown that the practical performance of MSMTFL-AT is less sensitive to the choice of $\tau^{(\ell)}$. In addition, we need to point out that the tuning parameter $\theta^{(\ell)} > 0$ is a key parameter, which typically decreases from a large value to a small value to  detect more correct nonzeros  from the gradually improved intermediate learning results as the iteration proceeds (as $\ell$ increases).

Intuitively, the ``first significant jump" rule works well partly because  the true nonzeros of  $\hat{t}^{(\ell)}$ are large in magnitude and small in number, while the false ones are large in number and small in magnitude. Therefore, the magnitudes of the true nonzeros are spread out, while those of the false ones are clustered.  The phenomenon of  ``first significant jump" was first observed in compressive sensing \cite{Wang2010} and  has been  
 observed  in other related sparse pattern recognition problems \cite{Wang2015a,Wang2015b}. While it is only a heuristic rule, it is still of practical importance, because a theoretically rigorous way to set the threshold value while keeping control of the false detections is still a challenging task in statistics \cite{Barbera14}.

\subsection{A Simple Demo}
We present a demo to show the effectiveness of the ``first significant jump" rule in multi-stage multi-task feature learning in Figure 1. We generate a sparse weight matrix $\bar{W}\in R^{200\times 20}$ and randomly set $90\%$ rows of it as zero vectors and $80\%$ elements of the remaining nonzero entries as zeros. We set $n=30$ and generate a data matrix $X\in R^{30\times 200}$ from the Gaussian distribution $N(0,1)$. The noise $\delta\in R^{30\times20}$ is sampled i.i.d from the Gaussian distribution $N(0,\sigma^2)$ with $\sigma =0.005$. The responses are computed as $Y=X\bar{W}+\delta$. The parameter estimation error is defined as $||\hat{W}-\bar{W}||_{2,1}$. 

In Figure 1,  $\bar{t}$ (a column vector) corresponds to the true weight matrix $\bar{W}$ and $\hat{t}^{(\ell)}$ corresponds to the learning weight matrix $\hat{W}$ in the $\ell$-th iteration. Subgraphs (a)-(d) plot $\hat{t}^{(1)}$, $\hat{t}^{(4)}$, $\hat{t}^{(7)}$, $\hat{t}^{(10)}$, respectively, in comparison with $\bar{t}$. From subgraph (a), it is clear that $\hat{t}^{(1)}$, which represents the solution of a Lasso-like model, contains a large number of false nonzeros and has a large recovery error, as expected.  As iteration proceeds, the intermediate learning result becomes more accurate. For example, $\hat{t}^{(4)}$ has a smaller error, and we can see that 
most of true nonzeros with large magnitude have been correctly identified. Then, $\hat{t}^{(7)}$ well matches the true learning matrix $\bar{t}$ even better, with only  a tiny number of false nonzero components. Finally, $\hat{t}^{(10)}$ exactly have the same nonzero components as the true $\bar{t}$, and the error is quite small. In short, Figure 1 shows that our proposed algorithm is insensitive to a small false number in $\hat{t}$ and has an attractive self-correction capacity.

\section{Numerical Experiments} \label{Sec:Experiments}
In this section, we demonstrate the better performance of the proposed MSMTFL-AT in terms of smaller recovery errors. We compare the MSMTFL-AT with a few competing multi-task feature learning algorithms: the MSMTFL, $\ell_{1}$-norm multi-task feature learning algorithm (Lasso),  $\ell_{2,1}$-norm multi-task feature learning algorithm (L2,1),  an efficient $\ell_{2,1}$-norm multi-task feature learning algorithm (Efficient-L2,1) \cite{Efficient} and a robust multi-task feature learning algorithm (rMTFL) \cite{rMTFL}.

\subsection{Synthetic Data Experiments}
As above, we denote the number of tasks as $m$ and each task has $n$ samples. The number of features is denoted as $d$. Each element of the data matrix $X_{i}\in R^{n\times d}$ for the $i$-th task is sampled i.i.d. from the Gaussian distribution $N(0,1)$ and each entry of the true weight $\bar{W}\in R^{d\times m}$ is sampled i.i.d. from the uniform distribution defined in the interval [-10, 10]. Here we  randomly set $90\%$ rows of $\bar{W}$ as zero vectors and $80\%$ elements of the remaining nonzero entries as zeros. Each entry of the noise $\delta_{i}\in R^{n}$ is sampled i.i.d. from the Gaussian distribution $N(0,\sigma^{2})$. The responses are computed as $y_{i}=X_{i}\bar{w}_{i}+\delta_{i}$.

The quality of the recovered  weight matrix is measured by the averaged  parameter estimation error $\|\hat{W}-\bar{W}\|_{2,1}$, which has a theoretical bound referring to \cite{Gong2013}. We present the averaged parameter estimation error $\|\hat{W}-\bar{W}\|_{2,1}$ vs. Stage $\ell$ in comparison with the MSMTFL and the MSMTFL-AT in Figure 2. It is clear that the parameter estimation errors of all tested algorithms decrease as the stage number $\ell$ increases, which shows the advantage of MSMTFL and MSMTFL-AT over the plain Lasso-like model ($\ell=1$). It is worth noting that the MSMTFL-AT is always superior to the MSMTFL under the same settings. Moreover, the parameter estimation error of our MEMTFL-AT decreases more rapidly and is more stable in most stages.

In order to show that the performance of MSMTFL-AT is insensitive to the settled value of the parameter $\tau$ to certain degree,  we depict the averaged parameter estimation error $\|\hat{W}-\bar{W}\|_{2,1}$ corresponding to  different $\tau$ values. The default value of $\tau$ is based on the heuristic formula (\ref{eq:tau}) and we also tried other values such as several times  the default value. The resultant recovery errors are listed in Table 1. We can see that, with different $\tau$ values,  our MSMTFL-AT achieves almost the same recovery accuracy and demonstrates the insensitiveness to the choice of the $\tau$ value.  Thus, in following  numerical experiments  we just select $\tau$ based on the heuristic formula (\ref{eq:tau}).

In Figure 3, we present the averaged parameter estimation error $\|\hat{W}-\bar{W}\|_{2,1}$ vs. lambda $\lambda$ in comparison with the MSMTFL with different $\theta$ ($\theta1=50m\lambda$, $\theta2=10m\lambda$, $\theta3=2m\lambda$ and $\theta4=0.4m\lambda$), the MSMTFL-AT, the $\ell_{1}$-norm multi-task feature learning algorithm (Lasso) and the $\ell_{2,1}$-norm multi-task feature learning algorithm (L2,1).  The settings of the parameters of these involved alternative algorithms  follow the same settings in their original papers. We compared the averaged parameter estimation errors of all the tested algorithms. As expected, the error of our MSMTFL-AT was the smallest among them. Specifically,  our proposed MSMTFL-AT significantly outperformed  the plain MSMTFL algorithm with the same settings suggested  in \cite{T.Zhang2009}.

In order to further demonstrate the advantage of the MSMTFL-AT, we compared it with  an efficient $\ell_{2,1}$-norm multi-task feature learning algorithm (Efficient-L2,1) proposed in \cite{Efficient} and a robust multi-task feature learning algorithm (rMTFL) proposed in \cite{rMTFL}. The Efficient-L2,1 solves  the $\ell_{2,1}$-norm regularized regression model  via  Nesterov's method. The rMTFL employs the accelerated gradient descent to efficiently solve the corresponding multi-task learning  problem, and has shown the scalability to large-size problems. 
The results are plotted  in Figure 4. We see that our MSMTFL-AT outperforms these two alternative  algorithms in terms of  achieving the smallest recovery error.

In short, these empirical results demonstrate the effectiveness of our proposed MSMTFL-AT. In particular, we have observed that (a) when noise level ($\sigma= 0.05$) is relatively large, the MSMTFL-AT also outperforms other tested algorithms and shows its robustness to the noise. (b) When $\lambda$ exceeds to a certain degree, the sparsity regularization weighs too much and the solutions $\hat{W}$ obtained by the involved algorithms will be too sparse. In such cases, the errors of all tested algorithms increase. Therefore, proper choice of $\lambda$ value is also very important. 

\subsection{Real-World Data Experiments}
We conduct a typical real-world data set, i.e. the Isolte data set (this data can be found at \url{www.zjucadcg.cn/dengcai/Data/data.html}) in this experiment. We aim to demonstrate the high efficiency of our algorithm in practical problems.  This Isolet data set is collected from $150$ speakers who speak the name of each English letter of the alphabet twice. Hence, we have $52$ training samples from each speaker. The speakers are grouped into $5$ subsets, each of which has  $30$ similar speakers. These subsets are referred to as isolet1, isolet2, isolet3, isolet4, and isolet5, respectively. Thus, there are $5$ tasks and each task corresponds to a subset. The $5$ tasks have $1560$, $1560$, $1560$, $1558$ and $1559$ samples, respectively (three samples historically are missing), where each sample has $617$ features and the response is the English letter label (1-26).

For our experiments, the letter labels are treated as the regression values for the Isolet sets. We randomly extract the training samples from each task with different training ratios ($15\%$, $20\%$ and $25\%$) and use the rest of the samples to form the test set. We evaluate four multi-task feature learning algorithms according to normalized mean squared error (nMSE) and averaged means squared error (aMSE), whose definitions  are as follows:
\begin{equation}
nMSE=n\frac{||\hat{y}-\bar{y}||_{F}^{2}}{||\hat{y}||_{1}\cdot||\bar{y}||_{1}},
\end{equation}
\begin{equation}
aMSE=\frac{||\hat{y}-\bar{y}||_{F}}{||\bar{y}||_{F}},
\end{equation}
where $\hat{y}$ is the predictive value from tested algorithms and $\bar{y}$ is the referent true value from the whole Isolet data set. Both nMSE and aMSE are commonly used in multi-task learning problems \cite{Gong2013, Y.Zhang2010}.

The experimental results are shown in Figure 5. We can see that our proposed MSMTFL-AT is superior to the Efficient-L2,1, rMTFL and MSMTFL algorithms in terms of achieving  the smallest nMSE and aMSE.  The MSMTFL-AT performs especially well even in the case of a small training ratio. All of the  experimental  results above suggest that the proposed MSMTFL-AT algorithm is a promising approach.

\section{Conclusions}
This paper proposes a non-convex multi-task feature learning formulation with an adaptive threshold parameter and introduces a corresponding MSMTFL-AT algorithm. The MSMTFL-AT is a combination of  an adaptive threshold learning via ``first significant jump" rule proposed in \cite{Wang2010} with the MSMTFL algorithm proposed in \cite{Gong2013}. The intuition is to refine the estimated threshold  by using intermediate solutions obtained from the recent stage. This alternative procedure between threshold value estimation and feature learning, leads to a gradually appropriate threshold value, and a gradually improved solution. The experimental results on both synthetic data and real-world data demonstrate the effectiveness of our proposed MSMTFL-AT in comparison with several state-of-the-art multi-task feature learning algorithms. In the future, we will give a theoretical analysis about the convergence of MSMTFL-AT. In addition, we expect to perform a rigorous analysis about why MSMTFL-AT could achieve a better practical performance than the original MSMTFL.

\section*{Acknowledgements}
This work was supported by the National Key Basic Research Program of China (No. 2015CB856000), the Natural Science Foundation of China (Nos. 11201054, 91330201)  and by the Fundamental Research Funds for the Central Universities( No. ZYGX2013Z005).  We would also like to thank the anonymous reviewers for their many constructive  suggestions, which have greatly improved this paper.

\begin{backmatter}

\begin{figure}[!htbp]
  \centering
   \subfigure[]{
   \label{fig:(a)}
   \includegraphics[width=0.35\textwidth]{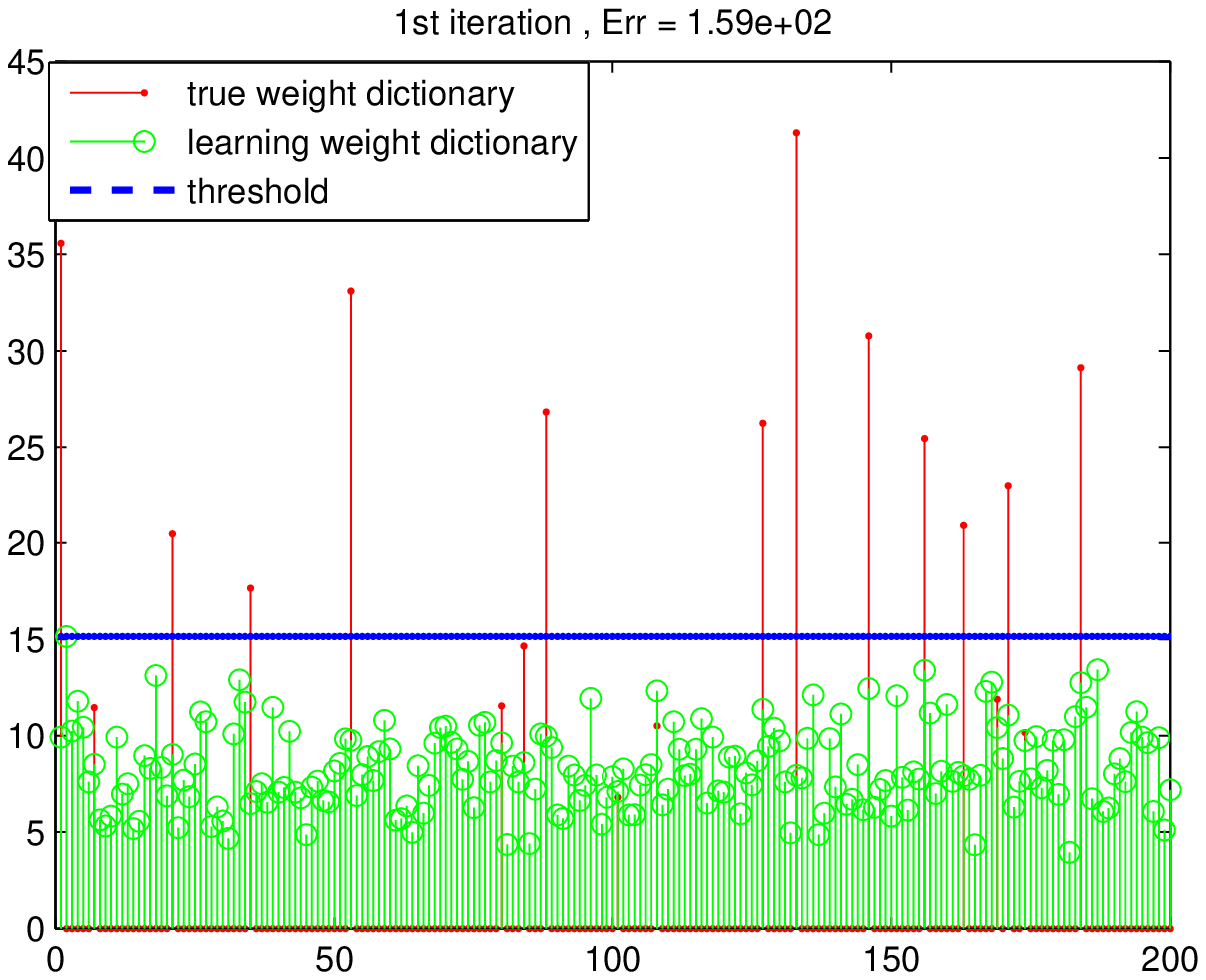}}
   \subfigure[]{
   \label{fig:(b)}
   \includegraphics[width=0.35\textwidth]{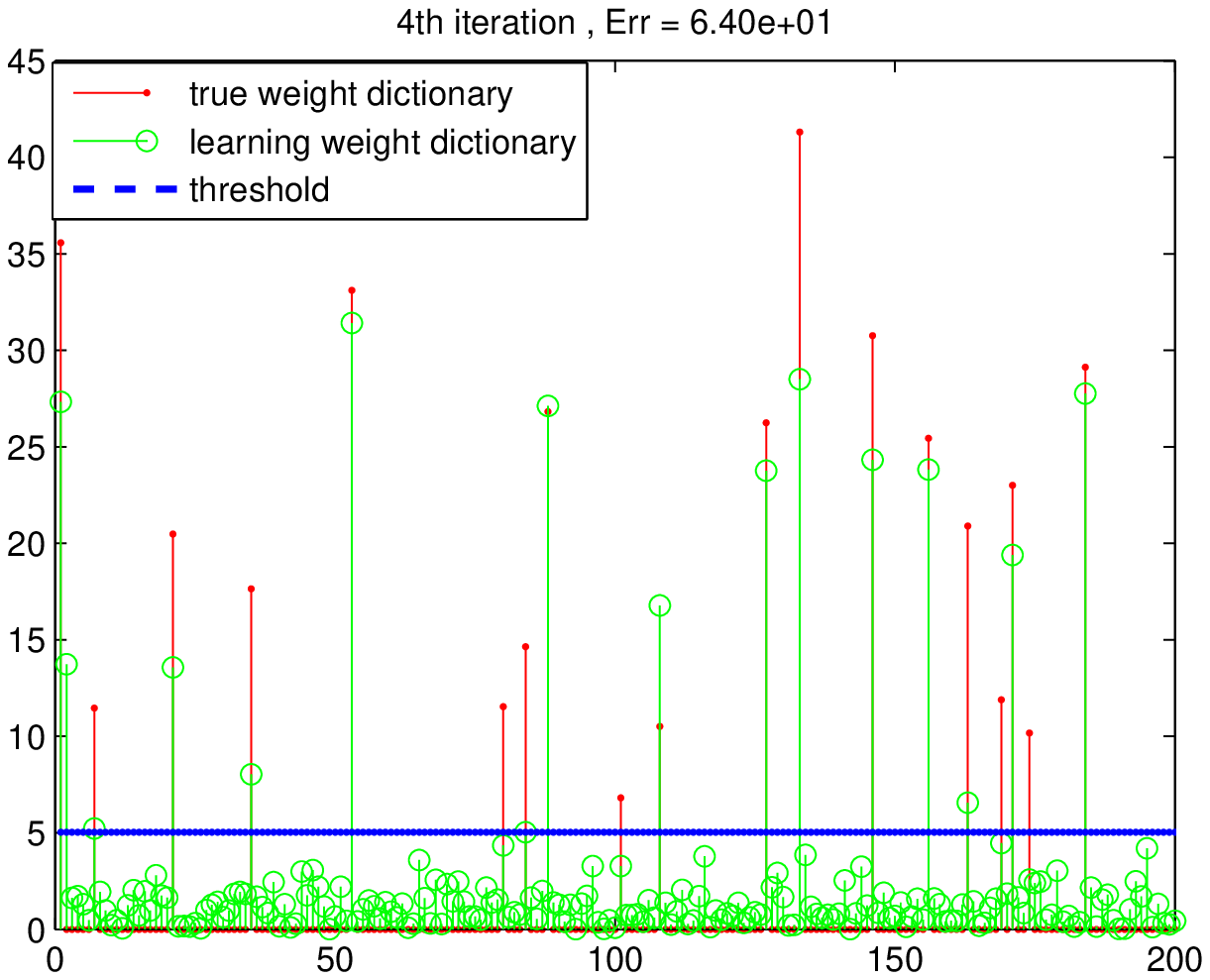}}
   \subfigure[]{
   \label{fig:(c)}
   \includegraphics[width=0.35\textwidth]{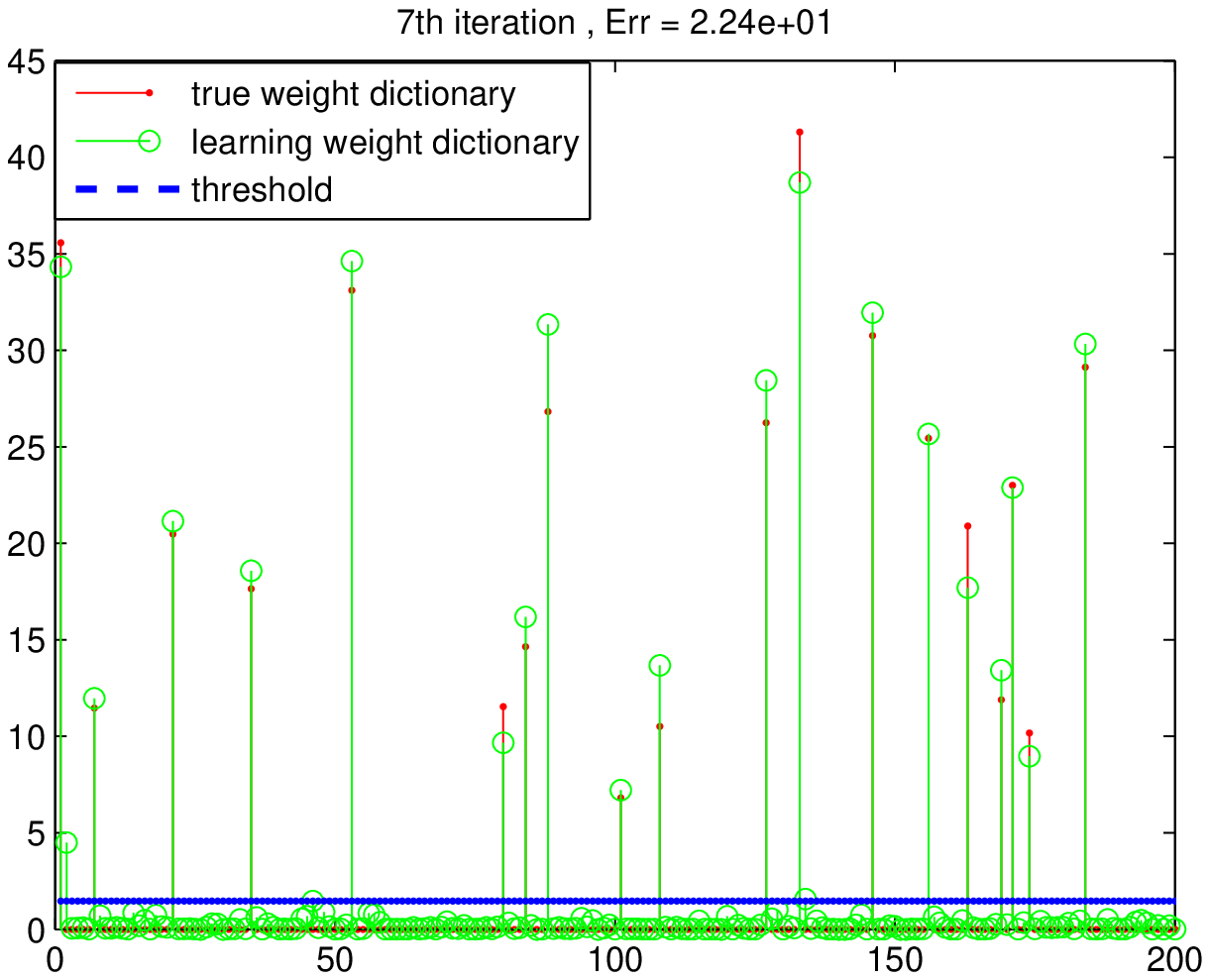}}
   \subfigure[]{
   \label{fig:(d)}
   \includegraphics[width=0.35\textwidth]{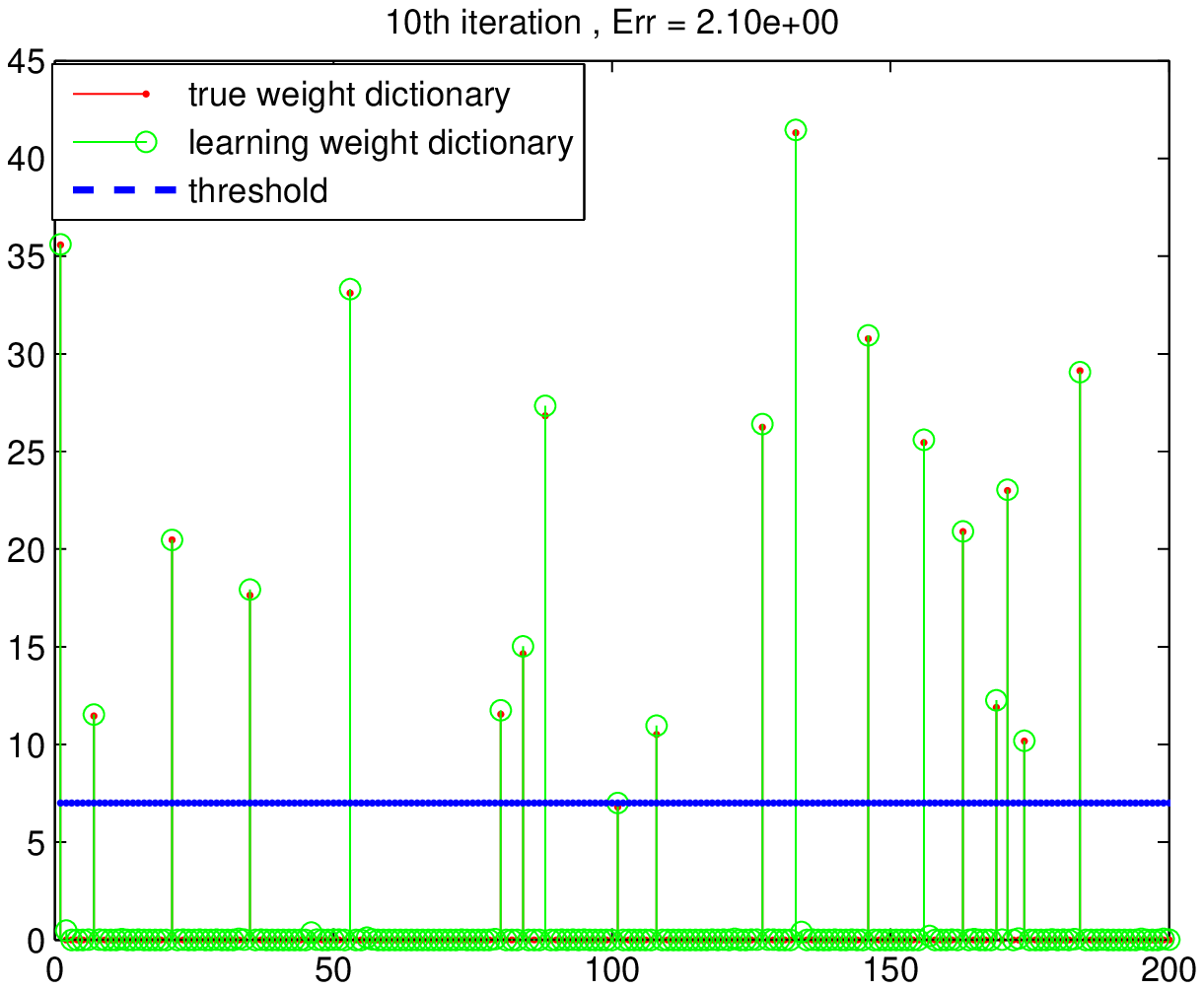}}
   \caption{a demo of "first significant jump" rule in multi-stage multi-task feature learning. (a)1st iteration, Error=1.59e+02, (b)4th iteration, Error=6.40e+01, (c)7th iteration, Error=2.24e+01, (d)10th iteration, Error=2.10e+00.}
\end{figure}

\begin{figure}[!htbp]
  \centering
   \subfigure[]{
   \label{fig:(a)}
   \includegraphics[width=0.3\textwidth]{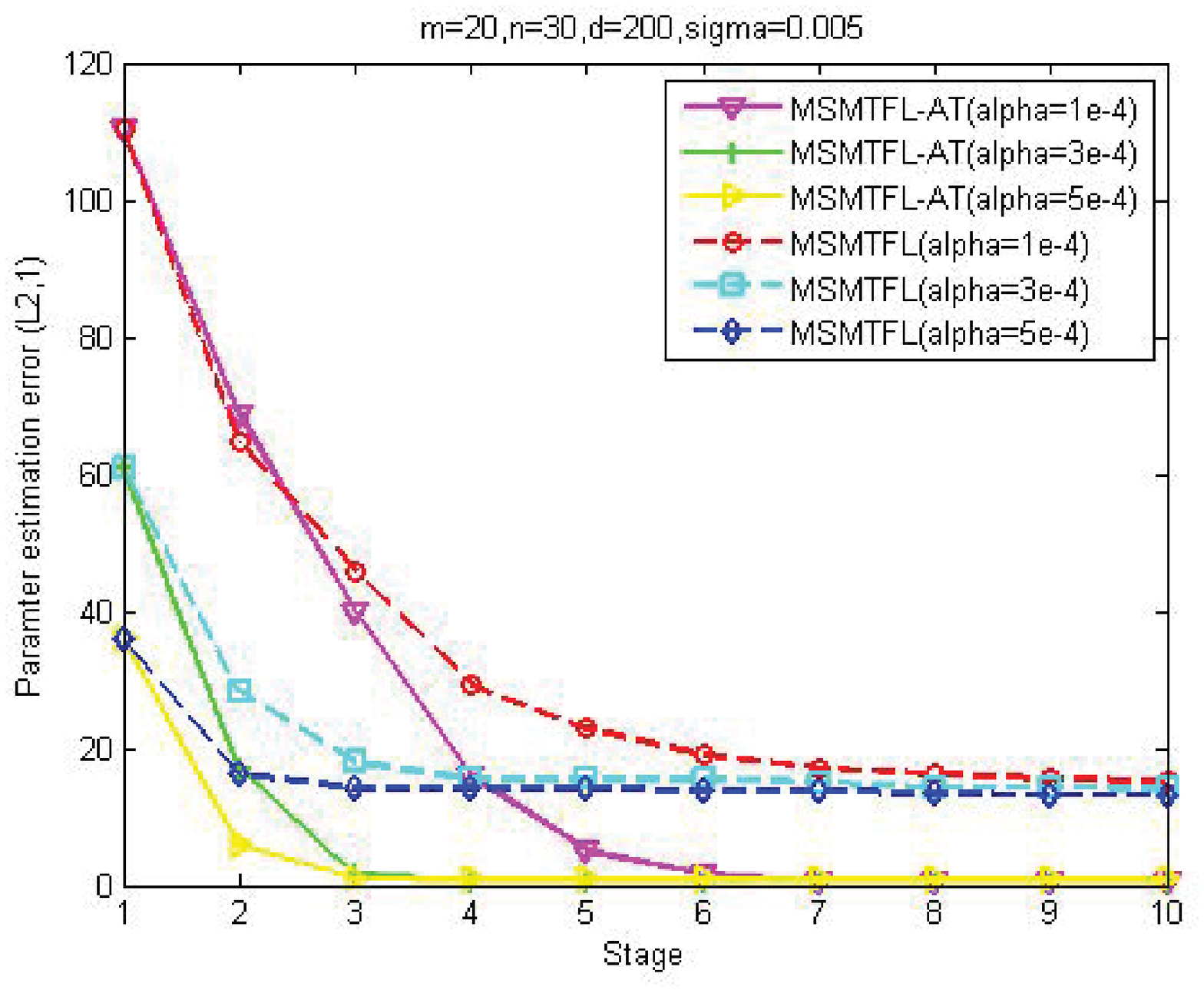}}
   \subfigure[]{
   \label{fig:(b)}
   \includegraphics[width=0.3\textwidth]{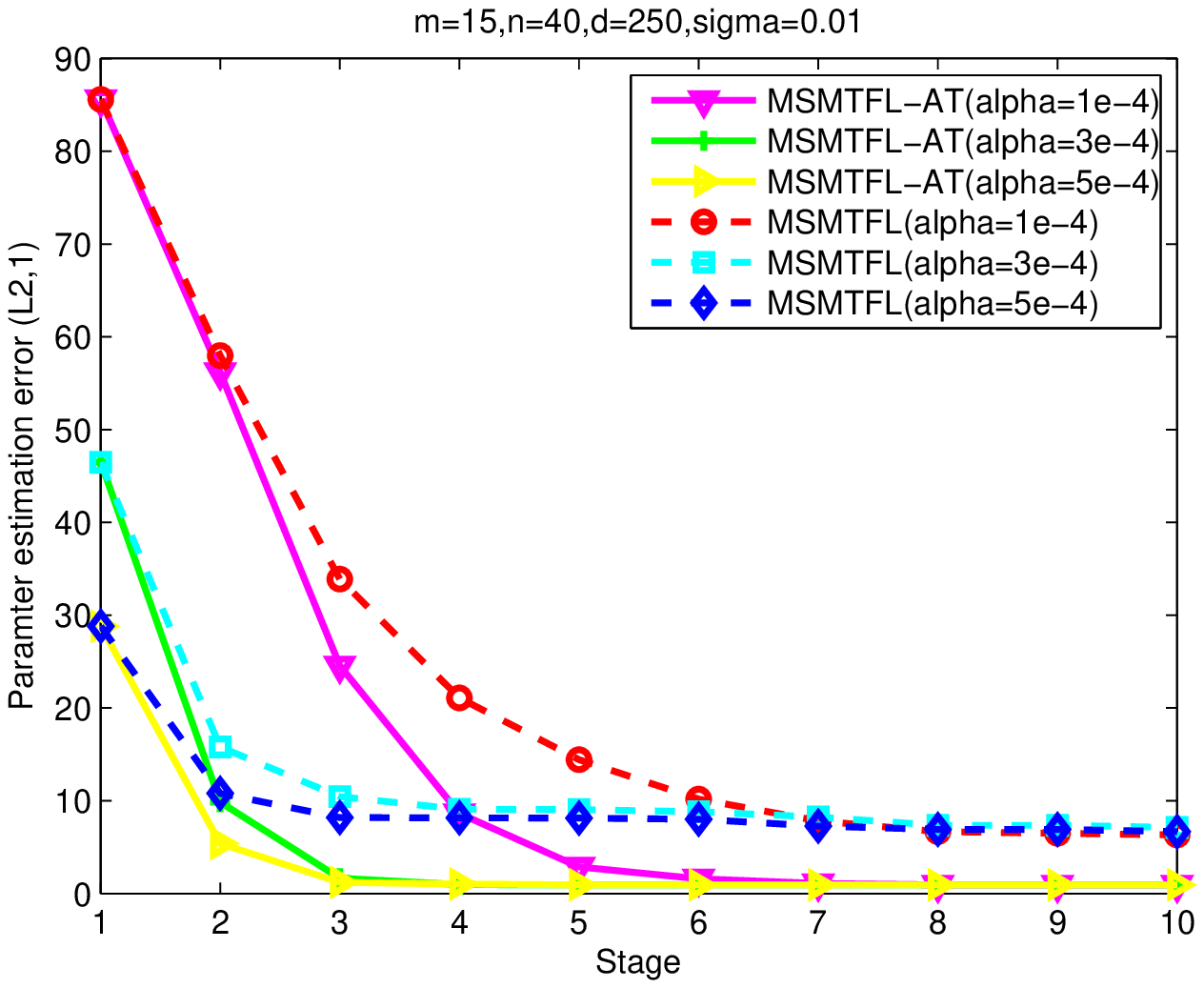}}
   \subfigure[]{
   \label{fig:(c)}
   \includegraphics[width=0.3\textwidth]{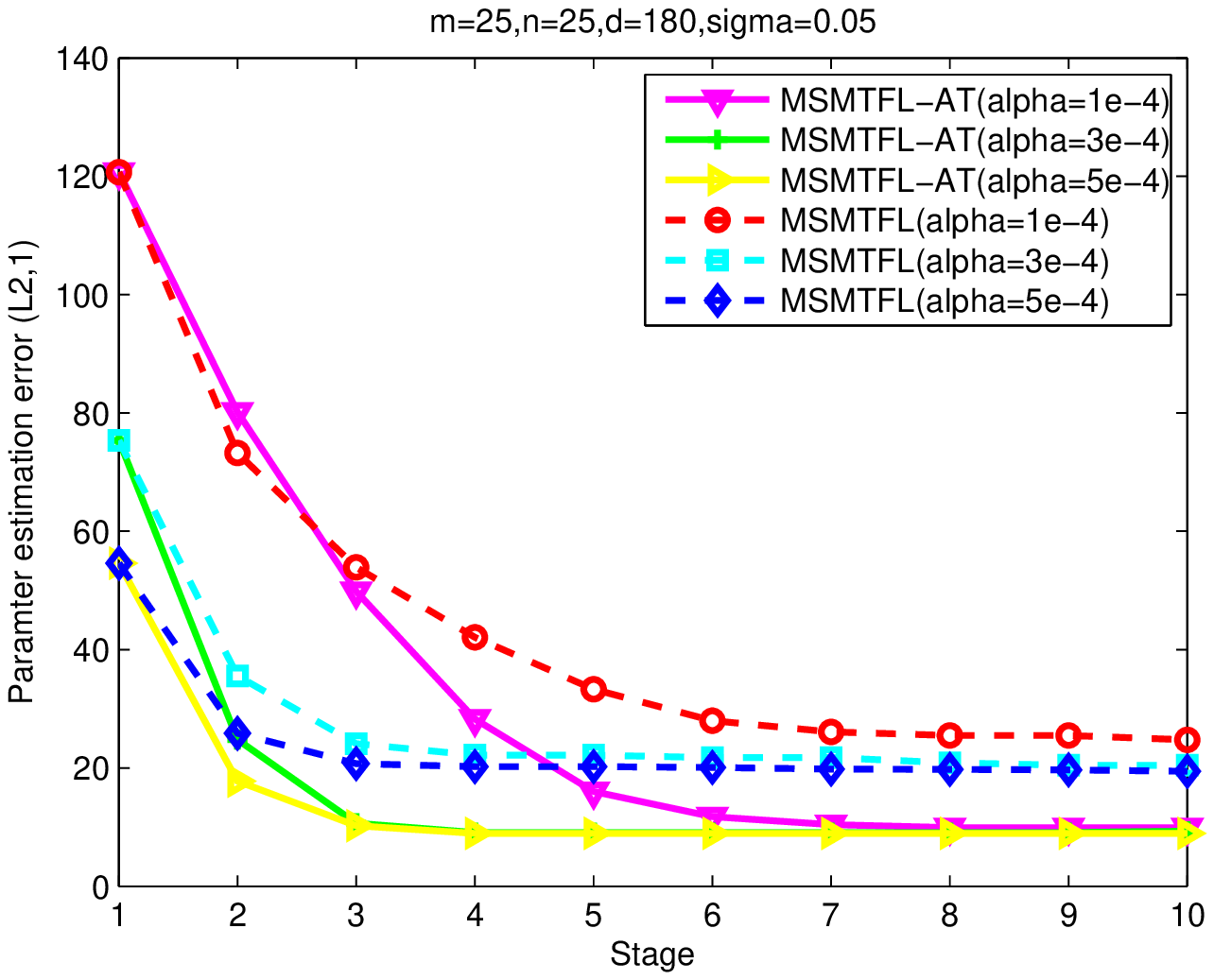}}
   \caption{Averaged parameter estimation error $\|\hat{W}-\bar{W}\|_{2,1}$ vs. Stage ($\ell$) plots for MSMTFL-AT and MSMTFL algorithms on the synthetic data set (averaged over 10 runs). We set $\lambda=\alpha\sqrt{ln(dm)/n}$, $\theta=50m\lambda$ referring to \cite{Gong2013}. (a)m=20, n=30, d=200, $\sigma$=0.005 (b)m=15, n=40, d=250, $\sigma$=0.01 (c)m=25, n=25, d=180, $\sigma$=0.05.}
\end{figure}

\begin{table}{
\centering
\begin{tabular}{|p{5cm}|p{2cm}|p{2cm}|p{2cm}|}
\hline 
  $\|\hat{W}-\bar{W}\|_{2,1}$ & 0.5$\tau$ & $\tau$ &  5$\tau$     \\
 \hline
    \cline{1-4}

      m=25, n=25, d=180, $\sigma$=0.05  & 8.2556 & 8.2725	& 8.8131 \\

      m=15, n=40, d=250, $\sigma$=0.01  & 0.9467 & 0.9458	& 1.0258\\

      m=20, n=30, d=200, $\sigma$=0.005 & 0.6442  & 0.6346   & 0.6785 \\

    \cline{1-4}
\hline
\end{tabular}}
\caption{\small Averaged parameter estimation error for the $\tau$ with different times .}
\end{table}

\begin{figure}[!htbp]
  \centering
   \subfigure[]{
   \label{fig:(a)}
   \includegraphics[width=0.3\textwidth]{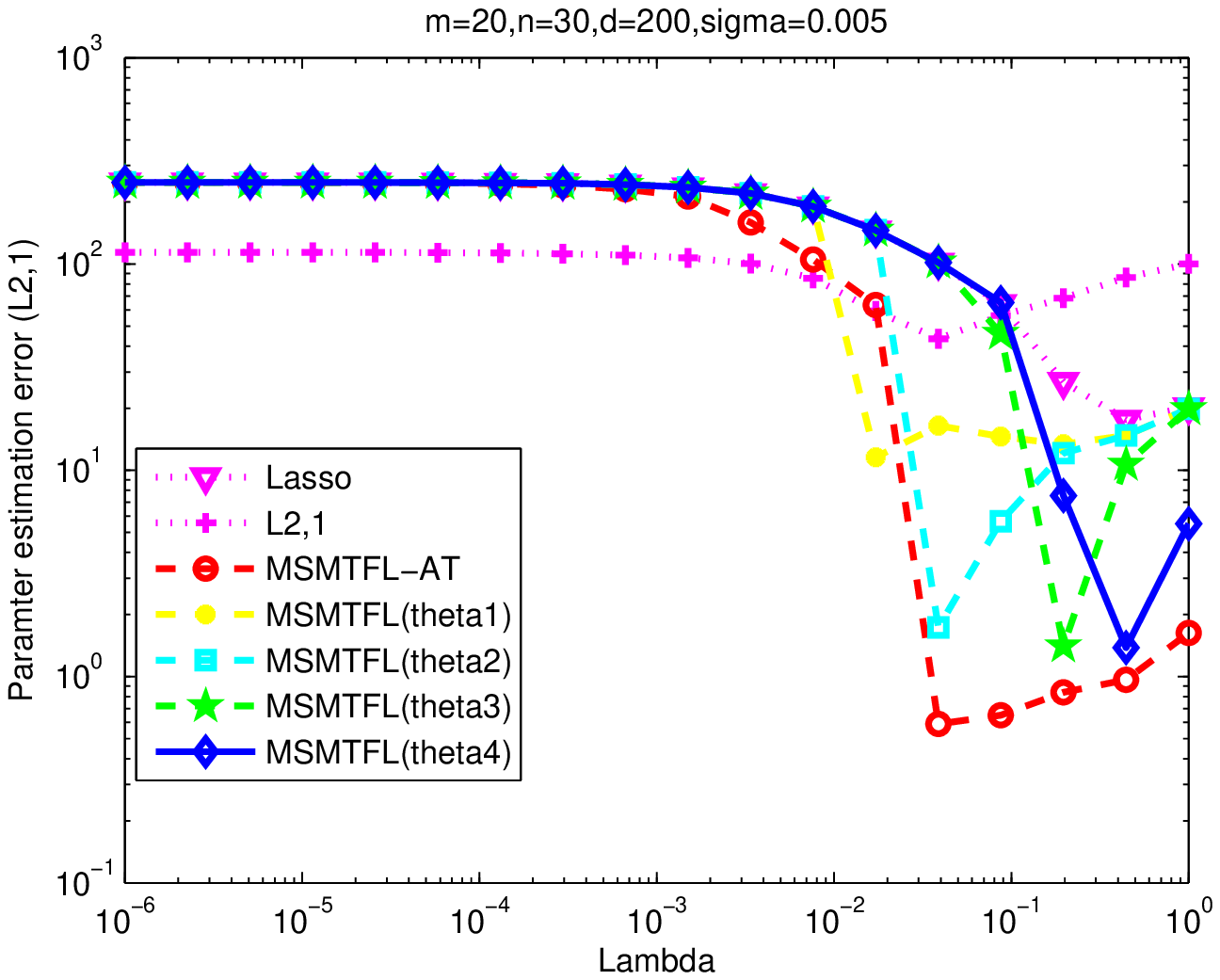}}
   \subfigure[]{
   \label{fig:(b)}
   \includegraphics[width=0.3\textwidth]{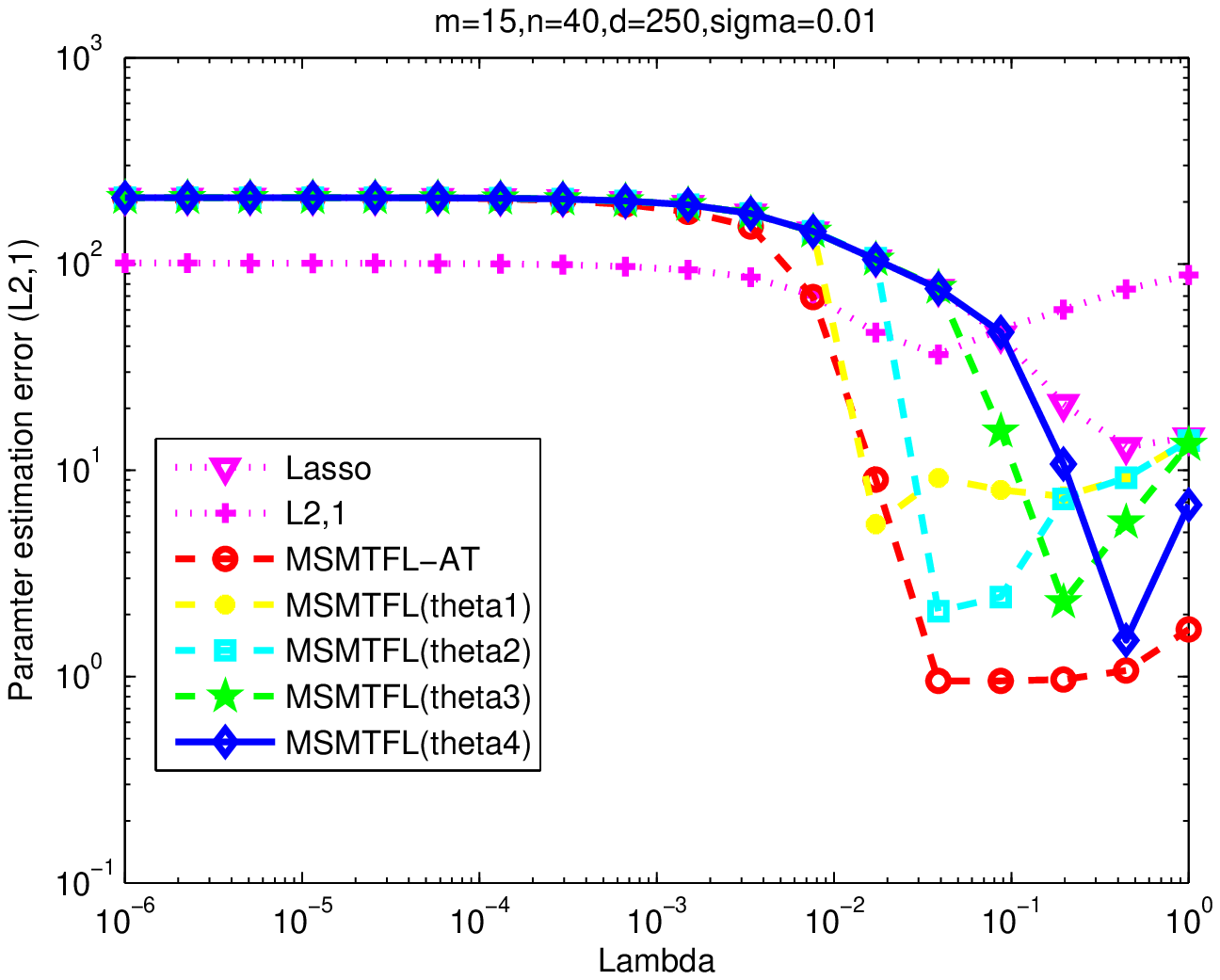}}
   \subfigure[]{
   \label{fig:(c)}
   \includegraphics[width=0.3\textwidth]{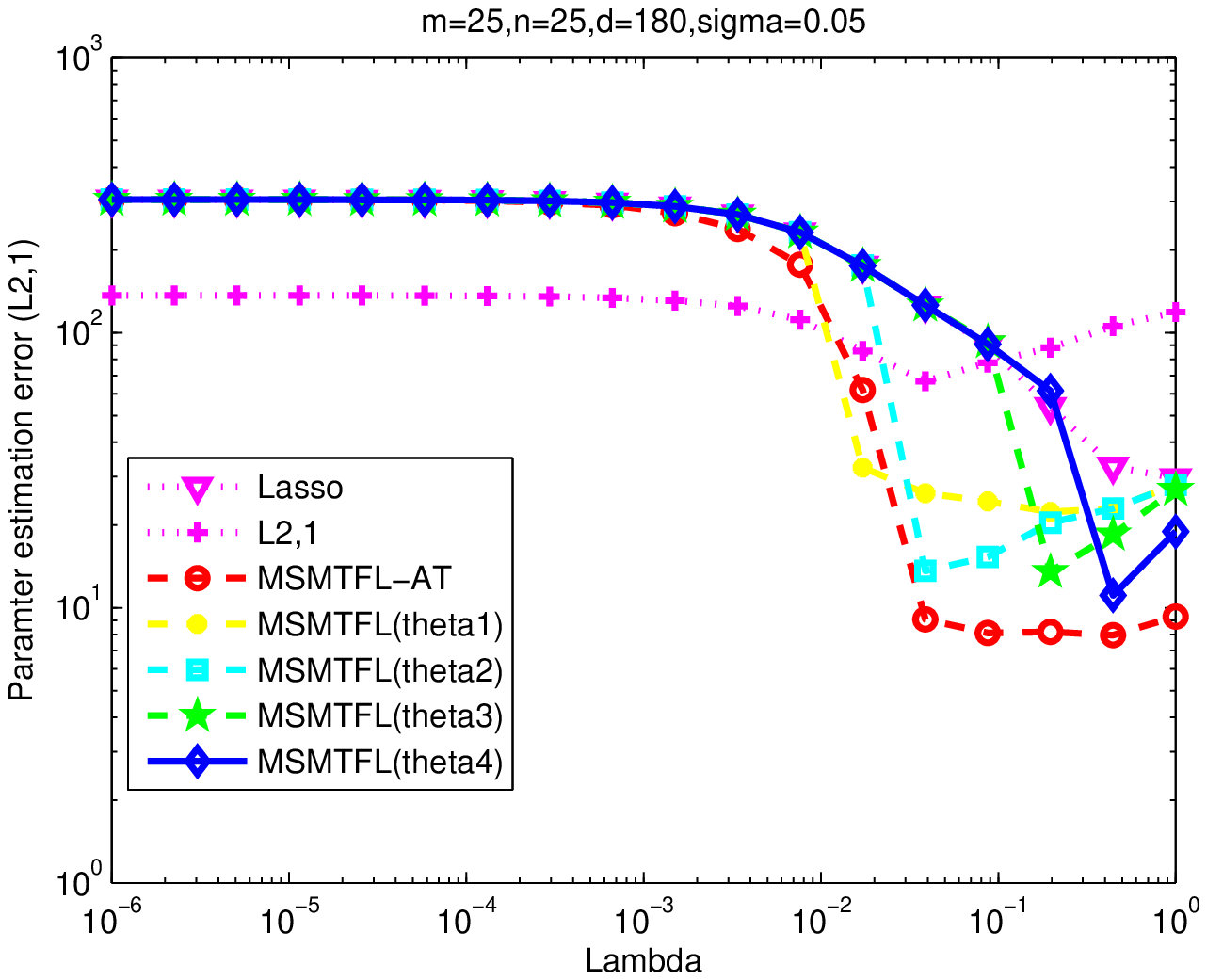}}
   \caption{Averaged parameter estimation error $\|\hat{W}-\bar{W}\|_{2,1}$ vs. Lambda ($\lambda$) plots for Lasso, L2,1, MSMTFL-AT and MSMTFL(theta1,2,3,4) algorithms on the synthetic data set (averaged over 10 runs). We set $\theta1=50m\lambda$, $\theta2=10m\lambda$, $\theta3=2m\lambda$ and $\theta4=0.4m\lambda$ for MSMTFL referring to \cite{Gong2013}.(a)m=20, n=30, d=200, $\sigma$=0.005 (b)m=15, n=40, d=250, $\sigma$=0.01 (c)m=25, n=25, d=180, $\sigma$=0.05.}
\end{figure}

\begin{figure}[!htbp]
  \centering
   \subfigure[]{
   \label{fig:(a)}
   \includegraphics[width=0.3\textwidth]{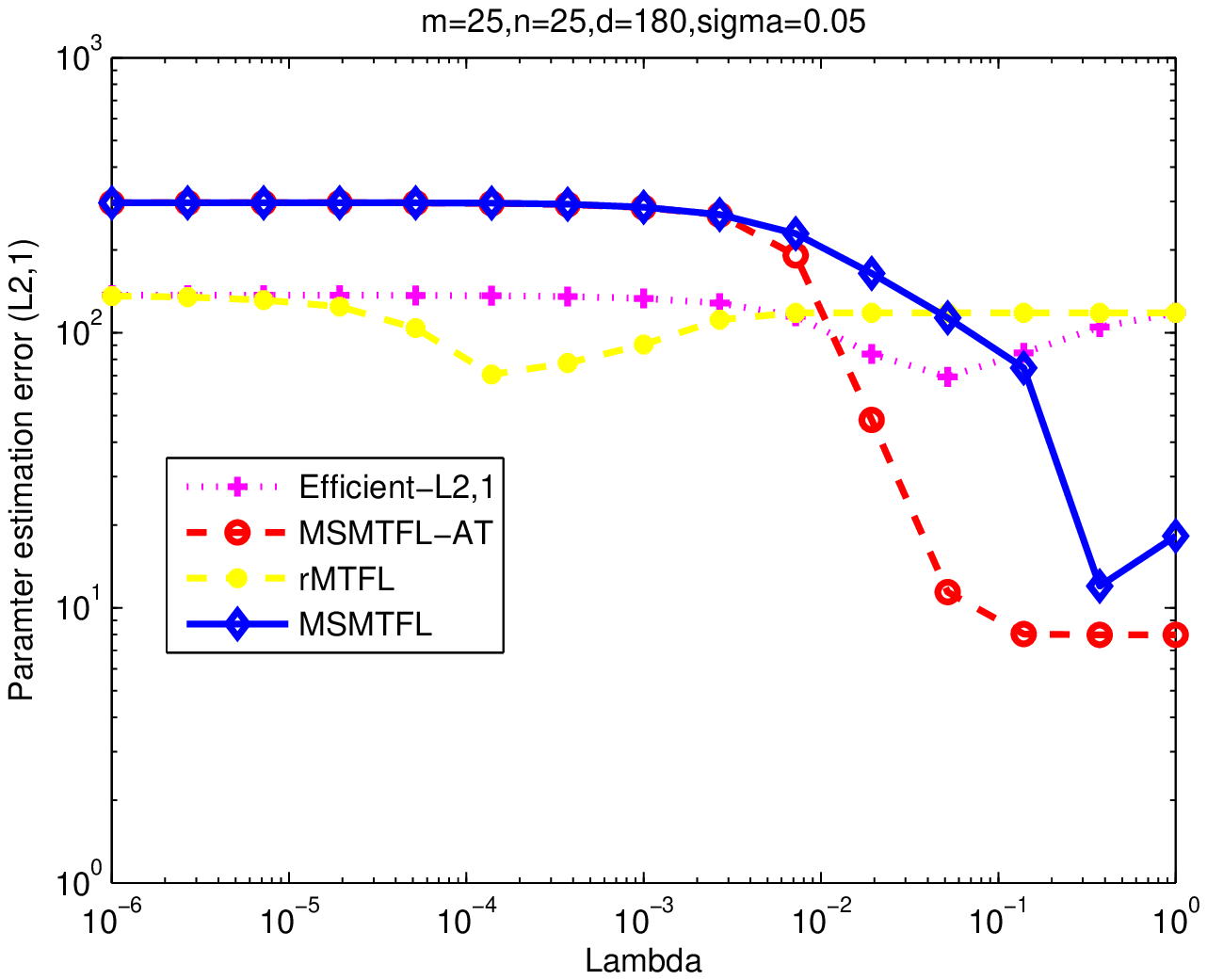}}
   \subfigure[]{
   \label{fig:(b)}
   \includegraphics[width=0.3\textwidth]{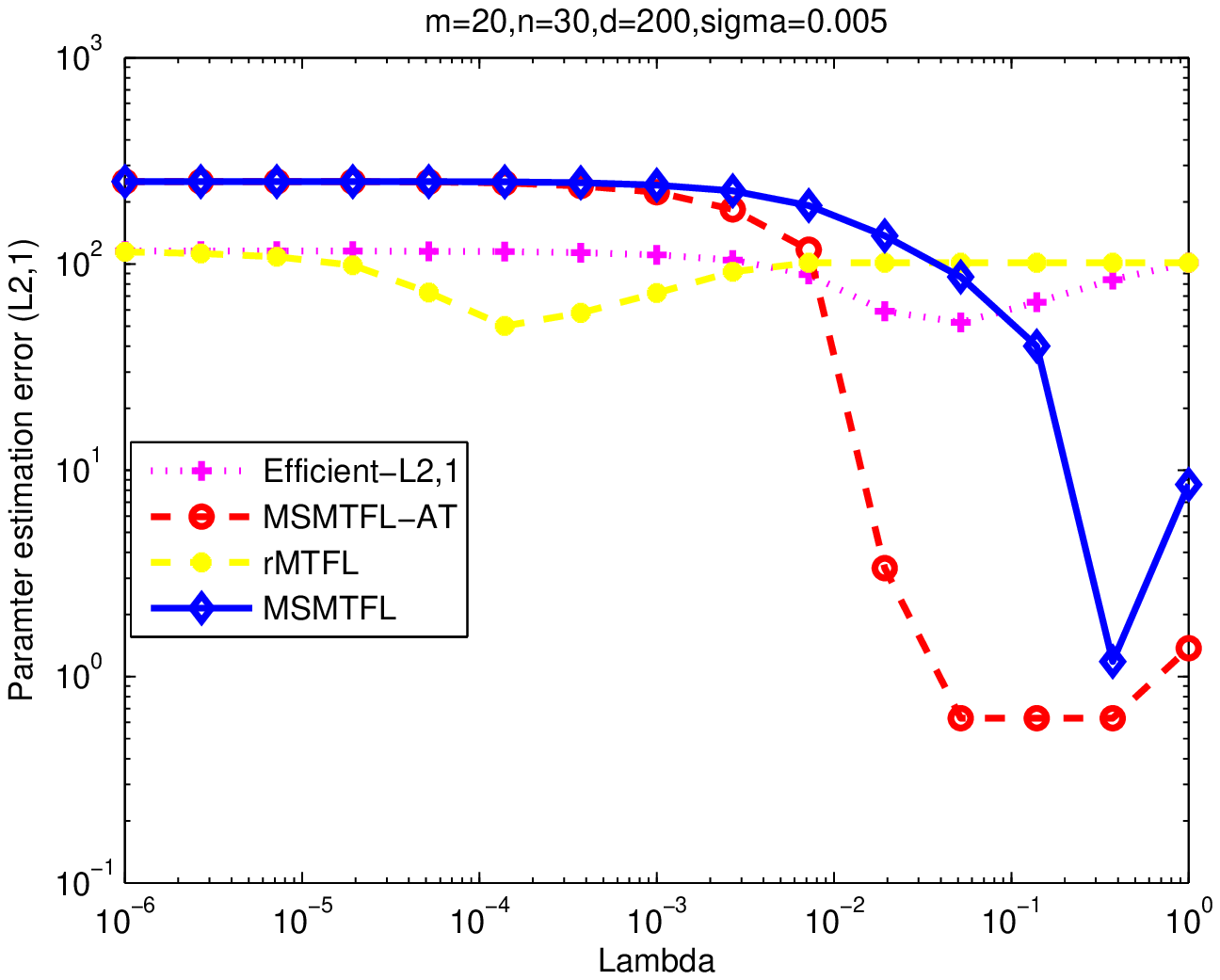}}
   \subfigure[]{
   \label{fig:(c)}
   \includegraphics[width=0.3\textwidth]{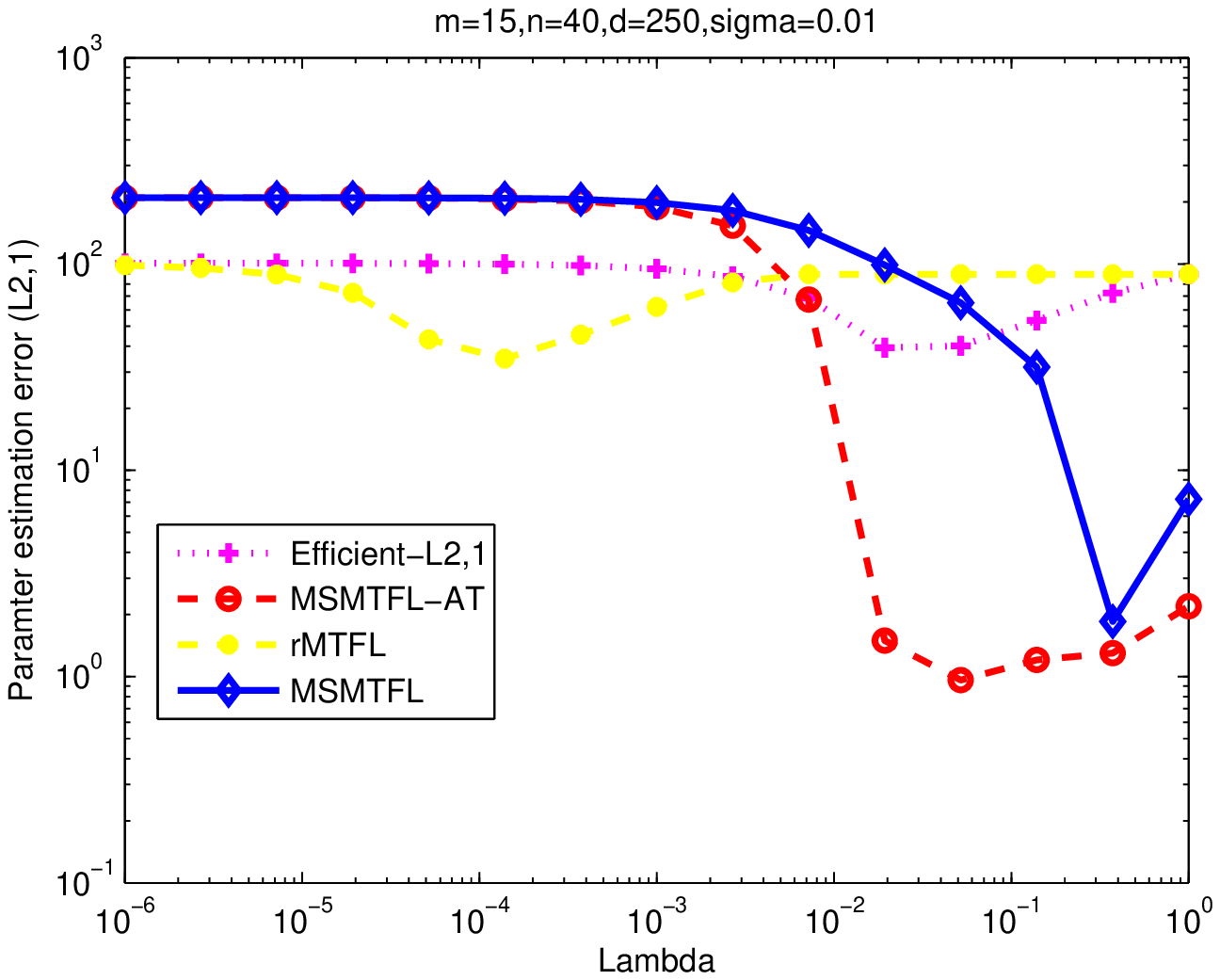}}
   \caption{Averaged parameter estimation error $\|\hat{W}-\bar{W}\|_{2,1}$ vs. Lambda ($\lambda$) plots for Efficient-L2,1, rMTFL, MSMTFL-AT and MSMTFL(theta4) algorithms on the synthetic data set (averaged over 10 runs).(a)m=25, n=25, d=180, $\sigma$=0.05 (b)m=20, n=30, d=200, $\sigma$=0.005 (c)m=15, n=40, d=250, $\sigma$=0.01.}
\end{figure}

\begin{figure}[!htbp]
  \centering
   \subfigure[]{
   \label{fig:(a)}
   \includegraphics[width=0.45\textwidth]{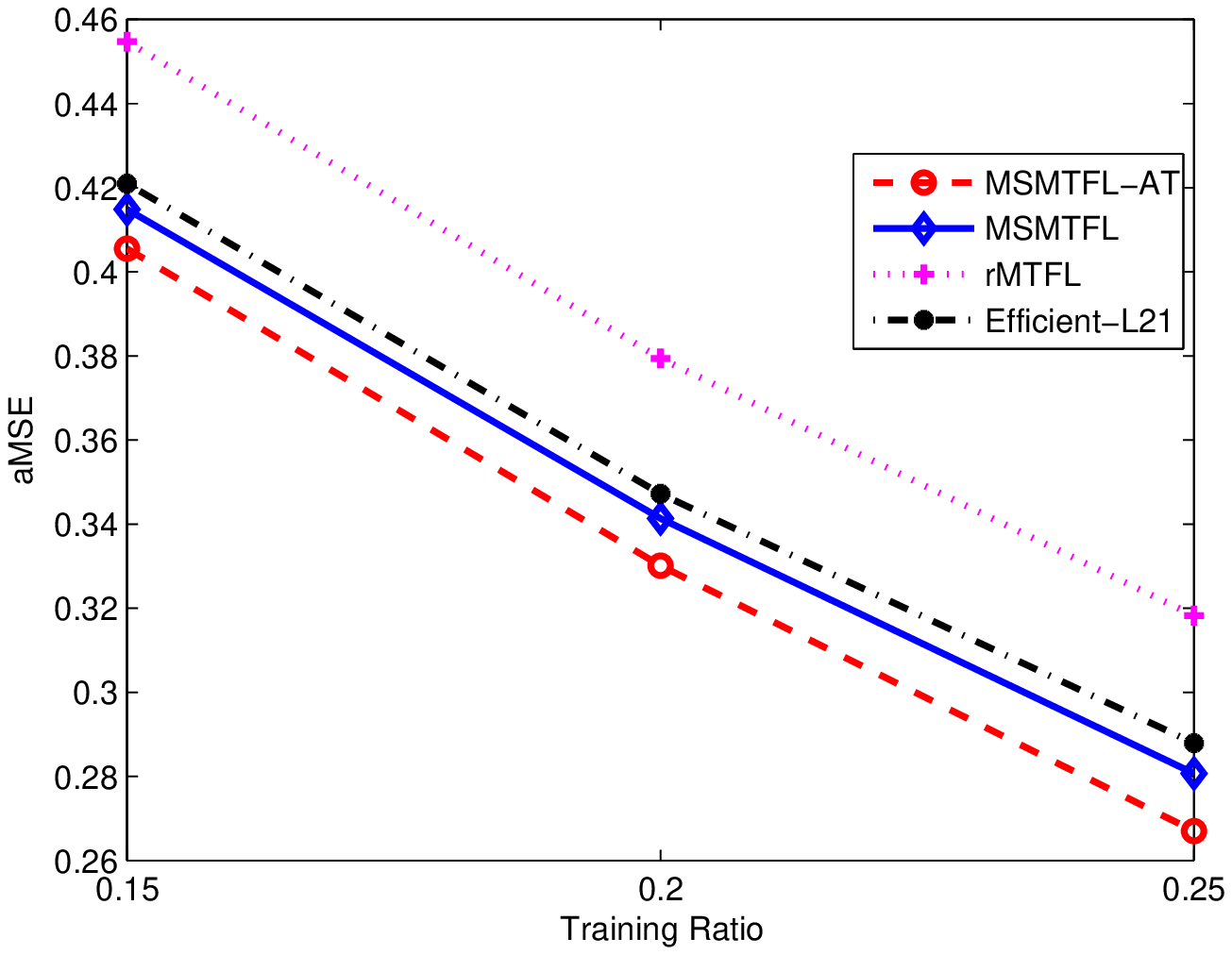}}
   \subfigure[]{
   \label{fig:(b)}
   \includegraphics[width=0.45\textwidth]{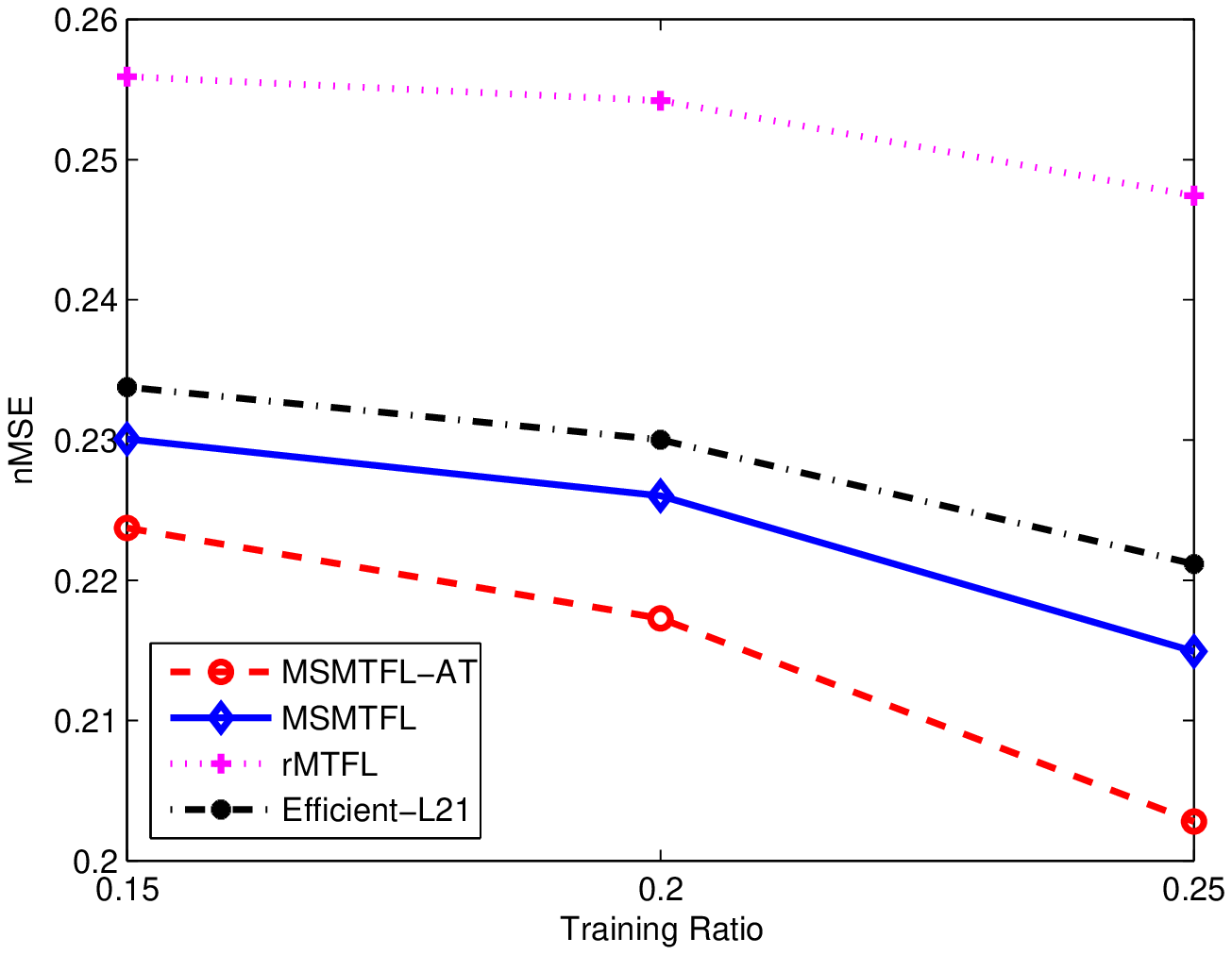}}
   \caption{(a) aMSE vs. training ratio plots on the Isolet data sets.(b) nMSE vs. training ratio plots on the Isolet data sets.}
\end{figure}
\end{backmatter}

\end{document}